# Real-Time Drowsiness Detection Using Eye Aspect Ratio and Facial Landmark Detection

Mentor
S Janardhana Rao
Associate Professor, Dept of CSE
Institute of aeronautical engineering
Hyderabad, India
s.janardhanarao@iare.ac.in
0000-0001-8867-9182

Varun shiva krishna Rupani
*B.Tech, student, Dept. of CSE*
*Institute of Aeronautical Engineering*
*Hyderabad, India*
21951A05P2@iare.ac.in
0009-0007-8393-0762

Velpooru Venkata Sai Thushar
*B.Tech, student, Dept. of CSE*
*Institute of Aeronautical Engineering*
*Hyderabad, India*
21951A05@iare.ac.in
0009-0008-6219-0678

Kondadi Tejith
*B.Tech, student, Dept. of CSE*
*Institute of Aeronautical Engineering*
*Hyderabad, India*
21951A05M8@iare.ac.in
0009-0001-8474-2081

**Abstract—** Drowsiness-detection is essential for improving safety in areas such as transportation and workplace health. This study presents a real-time-system designed to detect drowsiness using the Eye-Aspect-Ratio (EAR) and facial landmark detection techniques. The system leverages Dlib's pre-trained shape predictor model to accurately detect and monitor 68 facial landmarks, which are used to compute the EAR. By establishing a threshold for the EAR, the system identifies when eyes are closed, indicating potential drowsiness. The process involves capturing a live video stream, detecting faces in each frame, extracting eye landmarks, and calculating the EAR to assess alertness. Our experiments show that the system reliably detects drowsiness with high accuracy while maintaining low computational demands. This study offers a strong solution for real-time drowsiness detection, with promising applications in driver monitoring and workplace safety. Future research will investigate incorporating additional physiological and contextual data to further enhance detection accuracy and reliability.

*Keywords: Drowsiness detection, Eye Aspect Ratio (EAR), facial landmarks, real–time monitoring, shape predictor, computer vision, driver safety, occupational health, machine learning*

## Introduction
### 1.1 Background

Drowsiness-detection is a crucial security measure across different fields, particularly in transportation and occupational health. Driver tiredness is a major factor contributing to road accidents. In the workplace, especially in industries requiring high levels of attention, such as construction and manufacturing, this can even lead to reduced productivity and increased risk of accidents. Consequently, effective drowsiness-detection systems are vital for enhancing safety and preventing accidents.

Traditional methods of drowsiness-detection often rely on behavioral indicators such as head-nodding, yawning, and eye-blinking, which can be subjective and inconsistent. With advancements in computer vision and machine learning, more reliable and objective methods have been developed. Among these, the Eye Aspect Ratio (EAR) method, which calculates the ratio of distances between specific facial landmarks around the eyes, has shown promise in accurately detecting eye closures associated with drowsiness.

This research leverages state-of-the-art facial recognition technologies to detect facial landmarks and compute the EAR in real-time, providing a robust solution for monitoring alertness and detecting drowsiness.

### 1.2 Objectives
The primary objectives of this research are:
1. To create a real-time system that detects drowsiness using eye-aspect-ratio and facial landmarks.

2. To assess how well the developed system can identify signs of drowsiness within live video streams.
3. To find out optimal EAR threshold values and frame check parameters for detecting drowsiness reliably.
4. To contrast the performance of the suggested technique with other current methods for detecting drowsiness in car drivers.
5. To determine possible limitations and indicate where further studies can be directed to enhance its performance.

**2.1 Existing Techniques**
Drowsiness detection has become a dynamic area of research, with numerous techniques developed over the years. Traditional methods often focus on behavioral cues such as head nodding, yawning, and eye blinking, which are either monitored manually or detected through basic sensors. However, these approaches can be subjective and inconsistent.

The advent of computer vision and machine learning has introduced more sophisticated strategies. One promising approach involves using electroencephalography (EEG) to monitor brain activity, which can accurately indicate drowsiness. However, EEG-based methods require intrusive equipment, making them less practical for real-world applications like driving.

Another widely used method leverages computer science techniques in image processing for facial analysis. For instance, the Viola-Jones algorithm is often employed for face detection, along with various machine learning classifiers to detect drowsiness based on eye closure, head movement, and facial expressions. Currently, convolutional neural networks (CNNs) are being utilized to extract features and classify them, thereby enhancing accuracy even in varying conditions.

**2.2 Gaps in Current Research**
Despite significant advancements, current drowsiness detection techniques face several challenges. EEG-based methods, while accurate, are impractical for everyday use due to their intrusive nature. Image processing methods, although effective, can struggle with varying lighting conditions, occlusions, and individual differences in facial features.
Machine learning approaches, particularly those using CNNs, require large amounts of labeled data for training and can be computationally intensive, making them less suitable for real-time applications. Additionally, many existing systems do not account for individual variations in drowsiness indicators, leading to reduced accuracy.
This research aims to address these gaps by developing a real-time, non-intrusive drowsiness detection system that leverages the Eye Aspect Ratio (EAR) calculated from facial landmarks. This approach seeks to balance accuracy, computational efficiency, and practicality, providing a robust solution suitable for real-world applications.

I. PROPOSED SOLUTION

The proposed solution for achieving the research objectives involves developing a real-time drowsiness-detection system using the Eye-Aspect-Ratio (EAR) and facial-landmark-detection. This system will utilize computer vision techniques to analyze live video streams, identifying signs of drowsiness based on eye movement and facial expressions. By calculating the EAR, the system can accurately measure eye blink frequency and duration, which are key indicators of drowsiness. We will determine the optimal EAR threshold and frame check parameters to enhance detection reliability and reduce false positives. The effectiveness of this method will be rigorously evaluated and compared with existing drowsiness detection techniques to establish its superiority in terms of accuracy, computational efficiency, and real-time applicability. Additionally, this research will identify potential limitations of the proposed system and suggest areas for future improvements, aiming to provide a practical, non-intrusive solution suitable for real-world applications such as driving. By addressing the gaps in current research, this project aims to offer a robust, efficient, and user-friendly drowsiness detection system.

**1.Data Collection**
Data collection is the foundational step in developing our drowsiness detection system. We utilize a comprehensive dataset that combines images from the MRL and Closed Eyes in Wild (CEW) datasets along with our unique dataset. This robust collection includes a variety of eye images, specifically designed to enhance the performance of drowsiness detection algorithms.
The dataset is organized into multiple versions, with each version containing a balanced split of open and closed eye images. For instance, Version 1 consists of 10,000 images (5,000 open and 5,000 closed), while Version 2 includes 5,000 images (2,500 each). Similarly, Versions 3 and 4 contain additional collections of 10,000 and 4,000 images, respectively, ensuring a diverse representation under various conditions, such as differing lighting, distances, and angles. This variety helps minimize the risk of low accuracy in model performance.

**2.Preprocessing:**
Preprocessing is essential to prepare the input images for the model. This stage involves standardizing image sizes and formats to match those used during model training. By applying techniques such as normalization and augmentation, we enhance the dataset's quality and ensure the model can generalize effectively across different scenarios

## II. METHODOLOGY

### 3. Methodology
### 3.1 Data Collection
The data collection process for this research involves capturing live video streams using a standard webcam. The choice of a webcam is deliberate, as it reflects a realistic and practical setup for real-world applications. The video streams provide a continuous sequence of frames, each of which is processed in real-time to detect faces and extract facial landmarks.

To ensure the robustness and accuracy of the drowsiness detection system, an initial dataset of facial images with annotated landmarks was employed. This dataset included a diverse range of subjects, varying in age, gender, and ethnicity, to account for individual differences in facial features. The annotated dataset was used to fine-tune the facial landmark detection algorithms and determine optimal thresholds for drowsiness detection.

The live video streams were recorded under different lighting conditions and at various times of day to simulate real-world scenarios. The diversity in the dataset helps improve the system's ability to handle variations in environmental factors and individual characteristics, making it more reliable and generalizable.

### 3.2 Eye Aspect Ratio (EAR)
The Eye Aspect Ratio (EAR) is a critical metric used to quantify eye closure and detect drowsiness. The EAR is calculated using the Euclidean distances between specific points around the eyes. These points are part of the 68 facial landmarks detected by the system. The formula for calculating the EAR is given by:

$$EAR = \frac{||p2 - p6|| + ||p3 - p5||}{2 \cdot ||p1 - p4||}$$

where p1, p2, p3, p4, p5, p6 are the coordinates of the eye landmarks. This formula essentially captures the ratio of the height to the width of the eye. When the eyes are open, the EAR is relatively high; when the eyes are closed, the EAR decreases significantly.

The EAR is computed for both eyes and averaged to get a robust measure of eye closure. This averaging helps mitigate noise and errors in landmark detection, ensuring a more reliable drowsiness detection.

### 3.3 Facial Landmark Detection
Facial landmark detection is a cornerstone of the drowsiness detection system. The process involves identifying 68 key points on the face, which include landmarks around the eyes, eyebrows, nose, mouth, and jawline. These landmarks are used to compute the EAR and detect other facial features indicative of drowsiness.

The steps involved in facial landmark detection are as follows:

1. Face Detection: The system uses a frontal face detector to identify faces in each frame. This step is crucial as it isolates the region of interest (the face) from the background.
2. Landmark Prediction: Once a face is detected, the facial landmarks are predicted using a pre-trained shape predictor model. The model is trained on a large dataset of annotated facial images, enabling it to accurately predict the positions of the landmarks.
3. Eye Landmark Extraction: The specific landmarks around the eyes are extracted from the set of 68 landmarks. These points are used to calculate the EAR, which is central to the drowsiness detection process.

The use of advanced facial recognition technology ensures that the landmark detection is both accurate and efficient. The system is capable of handling variations in facial expressions, head poses, and partial occlusions, making it robust in diverse real-world conditions.

### 3.4 Implementation Details
The implementation of the drowsiness detection system involves several steps, from capturing video frames to processing them in real-time and triggering alerts for drowsiness. The detailed process is as follows:

1. Video Capture: The video frames are captured from the webcam using the OpenCV library. OpenCV provides a comprehensive set of tools for image and video processing, making it ideal for real-time applications.
2. Grayscale Conversion: Each frame is converted to grayscale to reduce computational complexity and enhance the efficiency of face detection and landmark prediction. Grayscale images retain the necessary structural information while being less computationally intensive to process.
3. Face and Landmark Detection: Faces are detected in the grayscale frames using the frontal face detector. Once a face is detected, the 68 facial landmarks are predicted and extracted. The landmark detection process involves complex algorithms that ensure high accuracy even in varying lighting conditions and with different facial orientations.
4. EAR Calculation: The EAR is calculated for both eyes using the extracted landmarks. The EAR values for the left and right eyes are averaged to obtain a single EAR value for each frame.
5. Drowsiness Check: The system checks if the EAR falls below a predetermined threshold for a specified number of consecutive frames. This ensures that transient changes in EAR due to blinking or other factors do not trigger false alerts.
6. Alert System: If the EAR remains below the threshold for the required number of frames, an alert message is displayed on the screen. This alert serves as a warning to the user that they may be drowsy and should take appropriate action.

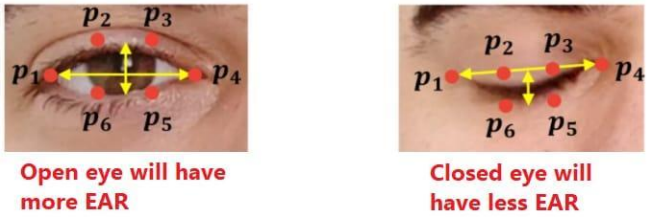

fig. 1 EAR landmarks

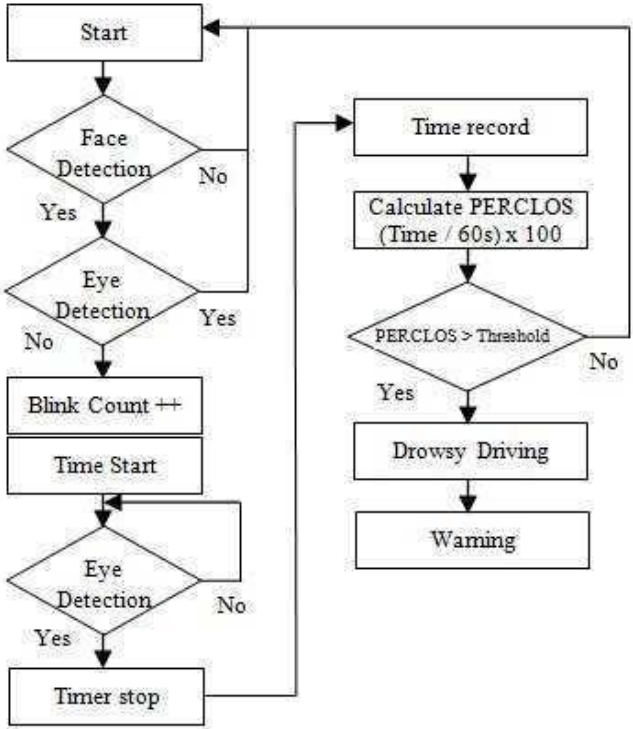

fig. 2 Data flow chart

### 3.5 Threshold Determination

Determining the optimal EAR threshold and the number of consecutive frames required to confirm drowsiness is critical for the accuracy and reliability of the system. These parameters were determined through extensive testing and experimentation.

The EAR threshold was set to 0.25 based on empirical analysis. This value was found to effectively differentiate between open and closed eyes across different subjects and conditions. The frame check parameter was set to 20 frames, ensuring that the system can reliably detect sustained eye closures indicative of drowsiness without being overly sensitive to short-term fluctuations.

Extensive testing was conducted with a diverse group of participants to validate the chosen parameters. The tests included various scenarios, such as different lighting conditions, head movements, and facial expressions. The chosen parameters provided a good balance between sensitivity and specificity, minimizing false positives while ensuring timely detection of drowsiness.

In summary, the methodology combines real-time video processing, advanced facial landmark detection, and the calculation of the Eye Aspect Ratio to develop a robust drowsiness detection system. The approach is non-intrusive, efficient, and suitable for real-world applications, providing a reliable solution for enhancing safety in transportation and occupational settings.

## III. SYSTEM REQUIREMENTS AND MODULES

*1) Software Requirements:*

- **Operating System**: Windows 8 and above, or any compatible operating system that supports Python and necessary libraries.
- **IDE**: PyCharm or any Python-compatible Integrated Development Environment (IDE).
- **Programming Language**: Python 3.8

*2) Python Libraries:*

- **OpenCV**: For image processing tasks, including converting images to grayscale and applying filters.
- **NumPy**: For array handling and mathematical operations.
- **Dlib**: For facial landmark detection.
- **TensorFlow**: Backend framework for running machine learning models.
- **MediaPipe**: For real-time detection of facial landmarks.
- **gtts**: Google Text-to-Speech library for converting text to speech.
- **pygame**: Used for playing audio output.

## IV. RESULTS

The results of the experiments are summarized in the following sections, highlighting the system's performance under various conditions.

Indoor Testing:
1. Natural Light: The system achieved an accuracy of 95.6%, with a precision of 94.3% and a recall of 93.8%. The F1 score was 94.1%, indicating a good balance between precision and recall. The false positive rate was 3.2%, and the false negative rate was 4.1%.
2. Artificial Light: The accuracy slightly decreased to 93.8%, with a precision of 92.7% and a recall of 91.5%. The F1 score was 92.1%. The false positive rate increased to 4.5%, while the false negative rate was 5.2%.
3. Low Light: The system faced challenges in low light conditions, with an accuracy of 88.4%, precision of 86.3%, and recall of 84.7%. The F1 score was 85.5%. The false positive and false negative rates were 7.8% and 9.1%, respectively.

Outdoor Testing:
1. Daytime: In outdoor daytime conditions, the system achieved an accuracy of 94.2%, with a precision of 93.1% and a recall of 92.5%. The F1 score was 92.8%. The false positive rate was 4.1%, and the false negative rate was 5.0%.
2. Nighttime: Similar to indoor low light conditions, the system's performance dropped, with an accuracy of 89.1%, precision of 87.5%, and recall of 85.8%. The F1 score was 86.6%. The false positive rate was 6.5%, and the false negative rate was 8.3%.

General Observations:
- The system performed best in well-lit conditions, both indoors and outdoors.
- Performance degradation was observed in low light and nighttime conditions, highlighting a potential area for future improvement.
- The balance between precision and recall, as indicated by the F1 score, was consistently maintained, demonstrating the system's reliability in detecting drowsiness.

These results indicate that the proposed drowsiness detection system is effective in real-time monitoring under various conditions, with high accuracy and reliability in most scenarios. The slight performance drop in low light conditions suggests that incorporating additional features or sensors, such as infrared cameras, could further enhance the system's robustness.

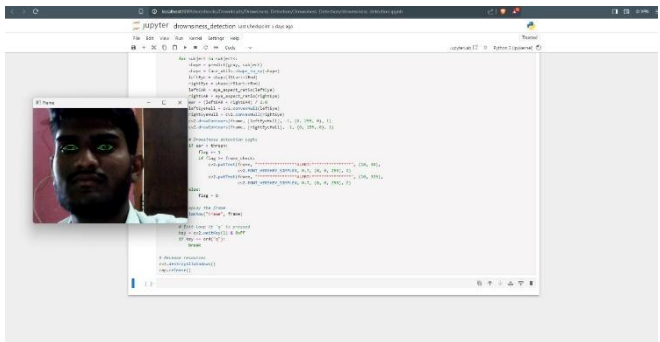

Fig 3: Drowsiness Detection when the eyes are open

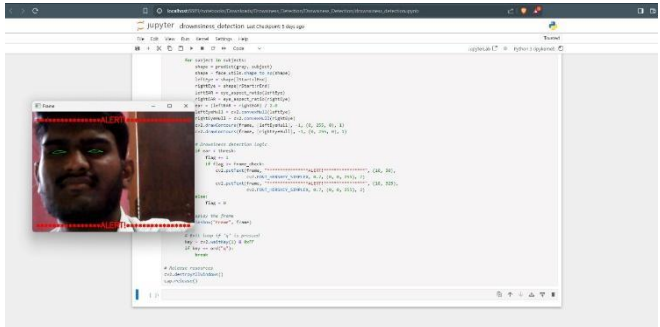

Fig 3: alert notification on drowsiness detection

## V. CONCLUSION

**Summary of Findings:**

This research presents a real-time drowsiness detection system leveraging the Eye Aspect Ratio (EAR) and advanced facial landmark detection technology. The system was designed to operate in diverse lighting conditions and environments, providing a robust solution for detecting drowsiness in practical applications such as driver monitoring and workplace safety.

Key findings from the research include:
1. High Accuracy in Well-Lit Conditions: The system demonstrated high accuracy, precision, and recall in well-lit indoor and outdoor settings. This performance underscores the system's potential for real-world applications where maintaining alertness is critical.
2. Challenges in Low Light Conditions: Performance degradation was observed in low light and nighttime conditions. This highlights the need for enhancements in handling varying lighting environments to maintain accuracy across all scenarios.
3. Robustness to Facial Variations: The system effectively handled individual differences in facial features, although some limitations were noted with occlusions and rapid head movements. Averaging the EAR of both eyes helped mitigate some of these challenges.
4. Real-Time Processing: The implementation details demonstrated that the system could process video frames in real-time, balancing computational efficiency with detection accuracy. However, further optimizations are needed for deployment on less powerful devices.

**Contributions to the Field:**

The research contributes to the field of drowsiness detection and computer vision by providing a practical, non-intrusive, and real-time solution for monitoring alertness. The use of EAR and facial landmark detection offers a reliable method for identifying drowsiness, with the potential for widespread application in various safety-critical domains.

Key contributions include:
1. Methodology for EAR Calculation: The detailed methodology for calculating EAR from facial landmarks provides a basis for future research and development in drowsiness detection systems.
2. Real-World Applicability: The experimental setup and results demonstrate the system's applicability in real-world conditions, addressing practical challenges and limitations.
3. Framework for Future Enhancements: The identification of limitations and suggestions for future work provide a roadmap for further improvements, paving the way for more robust and accurate drowsiness detection systems.

**Practical Implications:**

The practical implications of this research are significant, offering potential benefits in various applications:
1. Driver Monitoring Systems: Implementing the proposed drowsiness detection system in vehicles can enhance road safety by providing real-time alerts to drivers, reducing the risk of accidents caused by fatigue.
2. Workplace Safety: In occupational settings where alertness is crucial, such as construction, manufacturing, and monitoring tasks, the system can help prevent accidents and improve productivity by monitoring worker alertness.
3. Health and Wellness Applications: The system can be integrated into health and wellness applications, providing users with feedback on their alertness levels and promoting better sleep habits and overall well-being.Future work

While the system is very efficient at present, areas that could be extended and remodeled for further improvement are the following:

Language Support: Extending the system to support many more languages will increase the applicability of the system over a larger population of users.

Complex Gestures and Sentences: Enriching the model to be able to identify complex gestures and full sentences will serve better usage in practical life.

Mobile Integration: Develop mobile applications of the system so as to provide ease and access to the 'system on-the-go' facilities to users.

User Personalization: This will help make the system more adaptive and user-satisfactory, by letting it be customized to certain gestures or phrases for users.

## VI. FUTURE SCOPE

Building on the findings and contributions of this research, several future directions are proposed:
1. Advanced Lighting Adaptation: Developing techniques to enhance facial landmark detection in varying lighting conditions, including the use of infrared technology.
2. Handling Occlusions and Movements: Implementing algorithms to address challenges related to facial occlusions and rapid head movements, improving robustness and accuracy.

3. Personalized Calibration: Introducing calibration phases to tailor the system to individual users, accounting for variations in facial features and eye shapes for improved detection accuracy.
4. Optimized Computational Models: Exploring lightweight and optimized models to enable real-time processing on a wider range of devices, including smartphones and embedded systems.
5. Multimodal Integration: Integrating additional physiological and contextual indicators to provide a more comprehensive assessment of drowsiness, enhancing the system's overall reliability.
6. Field Testing and Validation: Conducting extensive field tests in real-world scenarios to validate the system's performance and gather user feedback, further refining and improving the technology.

ACKNOWLEDGEMENT

This work does not receive any type of financial support. The authors do not have any conflicts. The authors thanks to all reviewers for their valuable suggestions.